\documentclass[runningheads,orivec]{llncs}

\usepackage[T1]{fontenc}
\usepackage[utf8]{inputenc}

\usepackage{graphicx}
\usepackage{import}
\usepackage{amsmath}
\usepackage{subcaption}

\usepackage[T1]{fontenc}
\usepackage[utf8]{inputenc}
\usepackage{amsmath,amssymb,amsthm,bm,mathtools,physics}
\usepackage{placeins}
\usepackage{array}
\usepackage{algorithm}
\usepackage{algpseudocode,setspace}
\usepackage[english]{babel}
\usepackage{url,lineno,microtype}
\usepackage{mdframed}
\usepackage{upgreek}
\usepackage[disable]{todonotes}
\usepackage{soul}
\usepackage{enumitem}
\usepackage{changepage}
\usepackage{wrapfig}
\usepackage{multirow}
\usepackage{booktabs}
\newcommand{\KL}[2]{\operatorname{KL}\!\left[{#1}\,\|\,{#2}\right]}
\newcommand{\ent}[1]{\mathbb{H}\!\left[{#1}\right]}
\newcommand{\Ex}[2]{\mathbb{E}_{#1}\!\left[{#2}\right]}

\newmdtheoremenv{boxthm}{Theorem}
\newmdtheoremenv{boxlem}{Lemma}
\newmdtheoremenv{boxcor}{Corollary}
\newmdtheoremenv{boxdef}{Definition}

\usepackage{pgfplots}
\pgfplotsset{compat=1.17}
\usepackage{tikz}
\usetikzlibrary{intersections,positioning,shapes.geometric,backgrounds,calc,arrows,fit,patterns,decorations}
\usepgfplotslibrary{fillbetween,groupplots}
\pgfdeclarelayer{background}
\pgfdeclarelayer{main}
\pgfsetlayers{background,main}

\definecolor{beige}{RGB}{245,245,220}
\definecolor{darkgrey}{RGB}{75,75,75}
\definecolor{lightgrey}{RGB}{250,250,250}

\definecolor{thesischarcoal}{RGB}{51,51,51}
\definecolor{thesisamber}{RGB}{214,158,46}
\definecolor{thesisterracotta}{RGB}{192,96,64}
\definecolor{thesisdustyblue}{RGB}{96,127,160}
\definecolor{thesisteal}{RGB}{56,142,142}

\tikzstyle{smallbox} = [draw, minimum size=5.0mm]
\tikzstyle{box} = [draw, minimum size=7.0mm]
\tikzstyle{bigbox} = [draw, minimum size=10.0mm]
\tikzstyle{clamped} = [draw, fill=black, minimum size=0.15cm]

\usepackage{hyperref}

\newcommand{\refappx}[1]{\hyperref[#1]{Appendix~\ref*{#1}}}
\newcommand{\capsecref}[1]{\hyperref[#1]{Section~\ref*{#1}}}

\newif\ifanonymous
\anonymousfalse
 
\title{Sophisticated Policies from Epistemic Priors}

\ifanonymous
\author{Anonymous Authors}
\authorrunning{Anonymous Authors}
\institute{Anonymous Affiliation}
\else
\author{Wouter W. L. Nuijten\inst{1,2}\orcidID{0009-0007-0689-9300} \and
Bert de Vries\inst{1}\orcidID{0000-0003-0839-174X}}
\authorrunning{W. W. L. Nuijten \& B. de Vries}
\institute{Eindhoven University of Technology, 5612 AP Eindhoven, the Netherlands \and
Lazy Dynamics, Utrecht, the Netherlands}
\fi

\begin{document}

\maketitle

\begin{abstract}
    Sophisticated Inference is a variant of active inference often associated with recursive belief modeling and tree search. We argue that its central computational role is simpler: within a planning horizon, it makes active inference closed-loop by allowing future actions to depend on future states and observations. This closed-loop structure can be represented in the epistemic-prior variational free energy framework. Epistemic priors supply the active-inference objective, while a joint posterior over future states and actions supplies the state-contingent control structure.

    We evaluate this decomposition in the Reactivity Maze, a stochastic benchmark designed to separate epistemic incentive from inner-horizon closed-loop control. The comparison includes three variational objectives with the same state-action posterior family, an action-state factorized active inference objective, Sophisticated Inference, and standard Expected Free Energy planning. The results show that neither ingredient is sufficient on its own. Methods without an epistemic component do not seek information, while methods that prevent future actions from depending on future states cannot turn information into reliable goal-reaching. By contrast, both Sophisticated Inference and full-joint epistemic-prior active inference solve the environment by combining epistemic drive with closed-loop inference.

    These results show that the advantage associated with Sophisticated Inference need not be specific to tree search itself. It arises from the closed-loop form of active inference, and this form can be represented in epistemic-prior variational inference when the posterior keeps future actions dependent on future states.
\end{abstract}

\section{Introduction}\label{sec:introduction}

Active inference casts perception, learning, and action selection as variational inference in a generative model \cite{friston_freeenergy_2010,parr_active_2022}. In planning, an active inference agent evaluates possible futures before later observations have arrived. In stochastic settings, this requires more than scoring fixed action sequences: the planner must also represent how later actions would change under different future states or observations.

The Expected Free Energy (EFE) objective is central to active-inference planning because it combines instrumental and epistemic considerations in a single quantity \cite{friston_active_2015,dacosta_active_2020}. It says why preferred outcomes, ambiguity reduction, and information gain should matter. But an objective does not by itself determine the control structure used inside a planning rollout. A planner that evaluates a posterior over fixed action sequences $q(\bm{u})$ is open-loop within that rollout. A planner whose posterior admits conditional marginals such as $q(u_t \mid x_{t-1})$ is closed-loop within that rollout, because future actions can depend on the future states that are reached.

Sophisticated Inference is an active inference variant that addresses this limitation by branching over future contingencies \cite{friston_sophisticated_2021}. Its tree search represents possible future observations and belief states, and evaluates future actions conditionally at each branch. In this sense, Sophisticated Inference is already closed-loop active inference inside the planning horizon. This inner-horizon structure is separate from whether the whole planning computation is repeated after real observations arrive.

This paper asks whether the closed-loop structure used by Sophisticated Inference can be represented in the epistemic-prior variational free energy (VFE) framework. We argue that it can: epistemic priors supply the active-inference objective \cite{nuijten_message_2026}, and a joint posterior over future states and actions supplies the state-contingent control structure. This is a representational correspondence, not a claim of exact behavioral equivalence or of a specific inference algorithm. Epistemic-prior VFE therefore gives a variational way to represent the same closed-loop structure without identifying it with tree search.

This perspective leads to a concrete empirical claim. In the Reactivity Maze, the characteristic advantage associated with Sophisticated Inference need not depend on tree search itself, but on combining epistemic drive with posterior dependencies between future states and future actions. Epistemic components without those dependencies do not fully exploit informative cues, and state-action joint inference without epistemic drive has no reason to seek them out. Our contributions are as follows:
\begin{itemize}
    \item We clarify open-loop and closed-loop control within a single prospective rollout, distinguishing this inner-horizon structure from ordinary replanning after real observations.
    \item We separate the complementary roles of epistemic terms in the objective and posterior dependencies in the control representation, and show how epistemic-prior VFE instantiates their intersection.
    \item We evaluate this distinction in the Reactivity Maze, a benchmark designed to separate epistemic incentive from inner-horizon closed-loop control, and show that methods combining both ingredients reliably exploit informative cues and achieve the optimal outcome.
\end{itemize}

\capsecref{sec:background} reviews open-loop and closed-loop active inference and epistemic priors, \capsecref{sec:methodology} derives state-contingent action distributions from joint inference, and \capsecref{sec:evaluation}--\capsecref{sec:discussion} present the benchmark, results, and implications.

% =============================================================================
\section{Background \& Related Work}\label{sec:background}
% =============================================================================

% -----------------------------------------------------------------------------
\subsection{Open-Loop and Closed-Loop Control}\label{sec:planning-problem}
% -----------------------------------------------------------------------------

Sequential decision-making requires an agent to choose actions over time to achieve preferred outcomes. We focus on one prospective planning horizon before the next action is executed. Within that horizon, let $\bm{x}=x_{0:T}$ denote the current and future states and let $\bm{u}=u_{1:T}$ denote the future actions.

Open-loop control evaluates fixed action sequences, or distributions $q(\bm{u})$ over such sequences, without making later actions depend on the future states that may be reached. Closed-loop control evaluates future actions as conditional responses to future states or observations \cite{bertsekas_dynamic_2017,lavalle_planning_2006}. Equivalently, a closed-loop posterior can represent a state-dependent distribution over plan suffixes, such as $q(u_{t:T}\mid x_{t-1})$, with local conditionals $q(u_t\mid x_{t-1})$. In active inference papers, the word ``policy'' is often used for a distribution over action sequences; here it refers to these one-step conditional marginals.

The distinction matters whenever the predictive model assigns multiple possible future states to the same action sequence. For example, a short route with an obstacle blocking either the left or right passage may be preferable to a costly safe route, but only if the agent can later choose the open passage. An open-loop planner scores fixed continuations such as ``enter, then left'' and ``enter, then right'' separately; a closed-loop posterior can score ``enter, then choose left or right conditional on which passage is open''. That conditional continuation changes the value of entering the short route in the first place.
% -----------------------------------------------------------------------------
\subsection{Generative Models and Planning-as-Inference}\label{sec:generative-model}
% -----------------------------------------------------------------------------

Planning-as-Inference (PAI) formulates action selection as probabilistic inference in a predictive generative model \cite{attias_planning_2003,toussaint_probabilistic_2006}. For one prospective rollout, let $\bm{x}=x_{0:T}$, $\bm{y}=y_{1:T}$, and $\bm{u}=u_{1:T}$. A typical state-space model is
\begin{align}\label{eq:generative-model}
    p(\bm{y}, \bm{x}, \bm{u}, \theta) = p(x_0)\, p(\theta) \prod_{t=1}^T p(y_t \mid x_t, \theta)\, p(x_t \mid x_{t-1}, u_t, \theta)\, p(u_t)\,,
\end{align}
where $\theta$ are model parameters and all future rollout variables are unobserved. The prior $p(\bm{u})$ encodes admissible control sequences.

Within PAI, preferred outcomes are specified by a preference prior $\hat{p}(\bm{x})$ over state trajectories. This plays the role of a soft goal distribution and is closely related to exponentiated reward formulations in control and reinforcement learning \cite{levine_reinforcement_2018,todorov_linearlysolvable_2006,rawlik_stochastic_2012}. The planning problem is then to infer actions that steer the agent toward trajectories with high preference under $\hat{p}(\bm{x})$.

% -----------------------------------------------------------------------------
\subsection{Expected Free Energy and Standard Active Inference}\label{sec:efe}
% -----------------------------------------------------------------------------

Active inference is a PAI formulation grounded in the Free Energy Principle, balancing goal-directed and information-seeking behavior \cite{friston_action_2010,buckley_free_2017,dacosta_active_2020,dacosta_active_2024,parr_active_2022}. In the standard open-loop formulation, candidate action sequences are evaluated by the Expected Free Energy (EFE) \cite{friston_active_2015,dacosta_active_2020},
\begin{equation}\label{eq:G=r+a-n}
    G(\bm{u}) = \underbrace{\Ex{q}{\log \frac{q(\bm{x} \mid \bm{u})}{\hat{p}(\bm{x})}}}_{\text{risk}} + \underbrace{\Ex{q}{\log \frac{1}{q(\bm{y} \mid \bm{x}, \theta)}}}_{\text{ambiguity}} - \underbrace{\Ex{q}{\log \frac{q(\theta \mid \bm{y}, \bm{x})}{q(\theta \mid \bm{x})}}}_{\text{novelty}}\,,
\end{equation}
where expectations are with respect to $q(\bm{y}, \bm{x}, \theta \mid \bm{u})$. The terms measure risk, ambiguity, and novelty. Because $G(\bm{u})$ scores fixed future action sequences, it does not itself represent state-contingent future action selection. In stochastic settings, the planner must also represent how later actions would change after different future observations.

% -----------------------------------------------------------------------------
\subsection{Sophisticated Inference and Related Methods}\label{sec:si}
% -----------------------------------------------------------------------------

Sophisticated Inference (SI) \cite{friston_sophisticated_2021} addressed this limitation by branching over possible future observations and belief states, selecting future actions conditionally at each branch. SI is therefore already a closed-loop form of active inference within the planning horizon. Related methods, including Branching Time Active Inference \cite{champion_branching_2022} and Dynamic Programming EFE \cite{paul_predictive_2024,paul_efficient_2024}, share the same need to account for future adaptability. Our contribution is closest to variational and message-passing treatments of epistemic active inference \cite{palmieri_unifying_2022,parr_generalised_2019,koudahl_realising_2023,vandelaar_realizing_2025}: we show how the closed-loop structure used by SI can be represented in epistemic-prior VFE minimization.

% -----------------------------------------------------------------------------
\subsection{Epistemic Priors}\label{sec:epistemic-priors}
% -----------------------------------------------------------------------------

Recent work showed that the epistemic components of EFE can be absorbed into the generative model as epistemic priors, allowing active inference planning to be expressed as variational inference \cite{devries_expected_2025,nuijten_message_2026,nuijten_expected_2026}. For the same rollout, define
\begin{equation}\label{eq:VFE-for-planning}
    F[q] \triangleq \Ex{q(\bm{y}, \bm{x}, \bm{u}, \theta)}{\log \frac{q(\bm{y}, \bm{x}, \bm{u}, \theta)}{p(\bm{y}, \bm{x}, \bm{u}, \theta)\, \hat{p}(\bm{x})\, \tilde{p}(\bm{u})\, \tilde{p}(\bm{x})\, \tilde{p}(\bm{y}, \bm{x})}}\,,
\end{equation}
where the generative model is augmented by the preference prior $\hat{p}(\bm{x})$ and epistemic priors $\tilde{p}(\cdot)$. Choosing
\begin{subequations}\label{eq:epistemic-priors}
    \begin{align}
        \tilde{p}(\bm{u})         & \propto \exp\bigl(\ent{q(\bm{x} \mid \bm{u})}\bigr)\,, \label{eq:epistemic-prior-u}                                      \\
        \tilde{p}(\bm{x})         & \propto \exp\bigl(\Ex{q(\theta \mid \bm{x})}{-\ent{q(\bm{y} \mid \bm{x}, \theta)}}\bigr)\,, \label{eq:epistemic-prior-x} \\
        \tilde{p}(\bm{y}, \bm{x}) & \propto \exp\bigl(\KL{q(\theta \mid \bm{y}, \bm{x})}{q(\theta \mid \bm{x})}\bigr)\,, \label{eq:epistemic-prior-xy}
    \end{align}
\end{subequations}
gives
\begin{align}\label{eq:F=G+complexity}
    F[q] = \underbrace{\Ex{q(\bm{u})}{G(\bm{u})}}_{\substack{\text{expected plan} \\ \text{costs}}} + \underbrace{\Ex{q(\bm{y}, \bm{x}, \bm{u}, \theta)}{\log \frac{q(\bm{y}, \bm{x}, \bm{u}, \theta)}{p(\bm{y}, \bm{x}, \bm{u}, \theta)}}}_{\text{complexity}} + \text{const}\,.
\end{align}
where $G(\bm{u})$ is defined in \eqref{eq:G=r+a-n}. This encodes epistemic, ambiguity-reducing, and novelty-seeking behavior in a variational objective. Closed-loop control then depends on the variational family: if $q$ factorizes actions away from states, the planner remains open-loop; if $q$ retains dependencies such as $q(u_t\mid x_{t-1})$, the same VFE can support the structure that SI implements by tree search.

% =============================================================================
\section{From Joint Posteriors to Policies}\label{sec:methodology}
% =============================================================================

Epistemic priors specify why information-seeking behavior is valuable, while a joint posterior over states and actions specifies how contingent action selection is represented. Within one prospective rollout, closed-loop planning requires future action variables to remain dependent on future state variables. Outer-loop belief updating can still happen when real observations arrive, but it is not the mechanism isolated here.

% -----------------------------------------------------------------------------
\subsection{Policies from Joint Inference}\label{sec:policies-from-joint}
% -----------------------------------------------------------------------------

Throughout this section, $t=1,\ldots,T$ indexes positions inside the current planning horizon. Thus $q(u_t\mid x_{t-1})$ is not an update after real time has advanced; it is the conditional action distribution represented within the present rollout, before $x_{t-1}$ for $t>1$ has been observed.
Standard EFE evaluation cannot represent this dependency when it scores each candidate as a fixed action sequence with future actions independent of the future states that will actually be encountered (\capsecref{sec:efe}).

By contrast, VFE minimization over a joint posterior $q(\bm{y}, \bm{x}, \bm{u}, \theta)$ does not impose that independence.
When states and actions are inferred jointly, the posterior can represent statistical dependencies between them: different future states can support different future actions.
From the joint posterior, one can obtain $q(u_{t:T}\mid x_{t-1})$, a distribution over plan suffixes conditional on the state reached inside the rollout.
The conditional marginal $q(u_t\mid x_{t-1})$ is the one-step policy induced by that object.
This is the common core of the different narratives used in this paper: a closed-loop controller, a sophisticated agent, and a posterior without action-state independence constraints all refer to a model that can represent contingent action selection.

More generally, even without committing to any particular approximation family, VFE minimization gives an optimized joint posterior $q^*(\bm{y},\bm{x},\bm{u},\theta)$.
Marginalization and conditioning then expose the conditional policies encoded in that posterior:
\begin{equation}\label{eq:policy-marginalization}
    q^*(u_t | x_{t-1}) = \frac{q^*(u_t, x_{t-1})}{q^*(x_{t-1})}\,,
\end{equation}
where $q^*(u_t, x_{t-1})$ is obtained by summing the optimized joint posterior over all variables except $u_t$ and $x_{t-1}$.
The conditioning step in \eqref{eq:policy-marginalization} then yields the one-step policy.
This conditional is not added after optimization; it is read out from the dependencies already present in $q^*$.

The importance of \eqref{eq:policy-marginalization} is that it changes the valuation of the initial action, not only of later ones. A closed-loop posterior can favor $q(u_1\mid x_0)$ because it integrates over future branches together with future conditional responses. Receding-horizon control can still rerun an open-loop planner after each observation, but while choosing $u_1$ it scores only fixed continuations. The closed-loop object needed here is already present inside the current rollout.

% -----------------------------------------------------------------------------
\subsection{Decoupling Concept from Algorithm}\label{sec:decoupling}
% -----------------------------------------------------------------------------

The closed-loop requirement is not new; it is precisely the role of Sophisticated Inference~\cite{friston_sophisticated_2021}. SI achieves this through recursive tree search: at each future decision node, it evaluates candidate actions conditioned on the predicted belief state. Our contribution is to map that same requirement into epistemic-prior VFE, where branching is represented by posterior dependencies rather than explicit tree search. This is a representational correspondence, not an algorithmic identity.

Tree search is one way to represent the mapping from possible future states to future actions. A joint variational posterior is another, provided the approximation family does not impose $q(\bm{x},\bm{u})=q(\bm{x})q(\bm{u})$. The crucial property is not tree search itself, but the ability to represent state-contingent action selection.

The priors from~\eqref{eq:epistemic-priors} determine the epistemic objective; the joint posterior provides the capacity to act on it. Minimizing VFE of the augmented generative model therefore does not merely rank open-loop plans: when carried out over a joint posterior without action-state independence, it yields inner-horizon closed-loop control with the same qualitative advantage that Sophisticated Inference identified.

% =============================================================================
\section{Evaluation}\label{sec:evaluation}
% =============================================================================

\begin{figure}[t]
    \centering
    \begin{tikzpicture}[
    navstate/.style={circle, draw=thesischarcoal, thick, minimum size=8mm, inner sep=0pt, font=\small},
    goalstate/.style={navstate, very thick},
    special/.style={rectangle, rounded corners=3pt, draw=thesischarcoal, thick, minimum height=8mm, minimum width=14mm, inner sep=2pt, font=\small},
    flow/.style={-stealth, thin, thesischarcoal!40},
    arr/.style={-stealth, thick, thesischarcoal},
    darr/.style={-stealth, thick, dashed, thesischarcoal},
    influence/.style={-stealth, thick, dotted, thesischarcoal},
    note/.style={font=\scriptsize, align=center, text=thesischarcoal}
]

% --- Navigation ring (pentagon) ---
\def\R{1.5}
\node[goalstate, fill=thesisamber!20]      (s0) at ({90}:\R)     {$0$};
\node[goalstate, fill=thesisterracotta!20] (s1) at ({90-72}:\R)  {$1$};
\node[navstate]                       (s2) at ({90-144}:\R) {$2$};
\node[navstate]                       (s3) at ({90-216}:\R) {$3$};
\node[navstate]                       (s4) at ({90-288}:\R) {$4$};

% Clockwise navigation flow
\draw[flow] (s0) to[bend left=18] (s1);
\draw[flow] (s1) to[bend left=18] (s2);
\draw[flow] (s2) to[bend left=18] (s3);
\draw[flow] (s3) to[bend left=18] (s4);
\draw[flow] (s4) to[bend left=18] (s0);

% Goal labels (reward/loss colours as in the T-maze; theta selects which is rewarding)
\node[note, text=thesisamber!55!black]     at (90:\R+0.55) {$\theta{=}0$};
\node[note, text=thesisterracotta!70!black] at (30:\R+0.55) {$\theta{=}1$};

% Ring center annotation
\node[note] at (0, 0) {$(x^{(1)}_t {+} a_t)$\\[-1pt]$\bmod\; 5$};

% --- Safe sink ---
\node[special, fill=thesischarcoal!7] (sink) at (0, -3.5) {sink};
\node[note, below=1pt of sink] {$R = +0.33$};

% Stochastic fall arrow
\draw[darr] (0, -1.7) -- node[right, note, xshift=1pt] {prob.\ $1 - \tfrac{k}{4}$} (sink.north);

% --- Cue ---
\node[special, fill=thesisdustyblue!22] (cue) at (4.2, 0.3) {cue};
\node[note, above=2pt of cue] {reveals $\theta$};

% Action to cue (deterministic move)
\draw[arr] (s1.east) -- node[above, note] {$a_t {=} 7$} (cue.west);

% Uniform return from cue to a random ring state (stochastic)
\draw[darr] (cue.south west) to[out=210, in=-20] node[below, note, yshift=-1mm] {uniform return} (1.4, -0.7);

% --- Knob gauge ---
\def\knobX{-3.5}
\def\knobY{-0.4}
\def\knobW{0.45}
\def\knobH{2.4}

% Gauge outline
\draw[thick, thesischarcoal, rounded corners=2pt] (\knobX-\knobW/2, \knobY-\knobH/2) rectangle (\knobX+\knobW/2, \knobY+\knobH/2);

% Gauge fill segments
\foreach \k [evaluate=\k as \sat using {12+\k*20}] in {0,...,4} {
    \pgfmathsetmacro{\segBot}{\knobY - \knobH/2 + 0.08 + \k*0.44}
    \pgfmathsetmacro{\segTop}{\segBot + 0.36}
    \fill[thesisteal!\sat!white] (\knobX-\knobW/2+0.06, \segBot) rectangle (\knobX+\knobW/2-0.06, \segTop);
    \node[font=\tiny, text=thesischarcoal, anchor=west] at (\knobX+\knobW/2+0.06, {(\segBot+\segTop)/2}) {\k};
}

% Knob label
\node[note, anchor=south] at (\knobX, {\knobY+\knobH/2+0.18}) {knob $x_t^{(2)}$};

% Knob controls the stochastic fall probability (parameter influence)
\draw[influence] (\knobX+\knobW/2+0.3, \knobY-0.6) to[out=0, in=180] node[note, below, xshift=-14pt, yshift=-1pt] {controls} (-0.15, -2.45);

\end{tikzpicture}
    \caption{Schematic of the Reactivity Maze. Navigation states $0$--$4$ form a cyclic ring; the latent context $\theta \in \{0, 1\}$ selects whether state $0$ or state $1$ is the goal. The reactivity knob controls transition stochasticity: at low knob values, actions are likely to send the agent to the absorbing safe sink. The cue reveals $\theta$ but costs a timestep and carries a prior penalty.}
    \label{fig:reactivity-maze}
\end{figure}
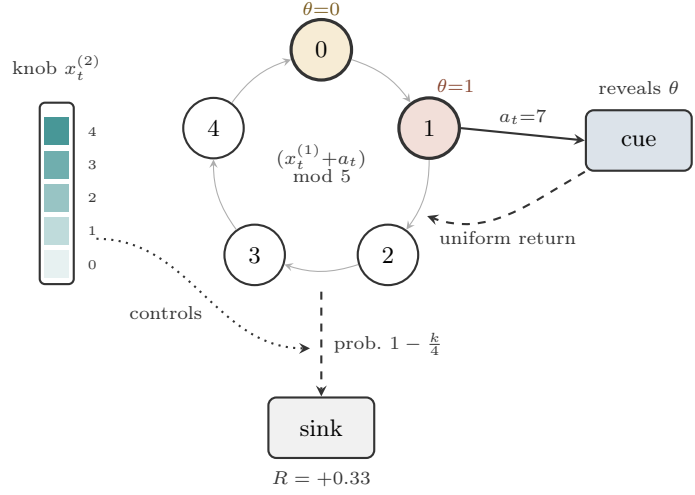

The evaluation uses an environment that separates the paper's two ingredients: epistemic incentive and inner-horizon closed-loop control. Standard T-maze settings do not suffice because deterministic transitions do not distinguish fixed action sequences from state-contingent future action choices. The Reactivity Maze, inspired by \cite[Appendix~I]{lazaro-gredilla_what_2024} and extended with latent uncertainty over the goal location, makes both contrasts explicit (\autoref{fig:reactivity-maze}).

The environment has two state factors.
The first factor $x_t^{(1)} \in \{0, 1, 2, 3, 4, 5, 6\}$ describes the agent's location: states $0$--$4$ are navigation states, state $5$ is an absorbing safe sink, and state $6$ is an instructional cue.
The second factor $x_t^{(2)} \in \{0, 1, 2, 3, 4\}$ is a reactivity knob that governs transition stochasticity.
There are eight actions: actions $0$--$4$ are navigation actions, actions $5$ and $6$ decrease and increase the knob, and action $7$ visits the cue.
A latent context parameter $\theta \in \{0, 1\}$ determines whether navigation state $0$ or $1$ is the rewarding goal.

This creates two strategies: a safe low-reactivity route that collapses into the absorbing sink, and a high-reactivity route that can reach the correct goal only by conditioning later actions on the state reached after visiting the cue and on the information gathered there.

The knob has deterministic dynamics: knob actions increment or decrement the knob value, clamped to $[0, 4]$.
The knob value controls the stochasticity of the location factor.
For navigation actions $0 \leq a_t \leq 4$ at non-special locations ($x_t^{(1)} \notin \{5, 6\}$):
\begin{equation}\label{eq:location-dynamics}
    x_{t+1}^{(1)} = \begin{cases}
        (x_t^{(1)} + a_t) \bmod 5 & \text{with probability } x_t^{(2)} / 4\,,     \\
        5 \;\text{(safe sink)}    & \text{with probability } 1 - x_t^{(2)} / 4\,.
    \end{cases}
\end{equation}
Knob actions ($a_t \in \{5, 6\}$) follow the same stochastic pattern: with probability $x_t^{(2)}/4$ the agent remains at its current location, otherwise it falls to the safe sink.
Action $7$ deterministically moves the agent to the cue location.
At the cue, any action returns the agent to a uniformly random navigation state.
The safe sink is absorbing.

At the cue location, the agent receives an observation that deterministically reveals $\theta$; at all other locations, observations are uninformative. After visiting the cue the agent resumes from a random navigation state, so exploiting the cue requires both learning the goal and adapting to the post-cue state.
The action prior assigns a lower probability to the cue action ($p(a_t = 7) \propto 1/\epsilon$ with $\epsilon = 1.1$), making information-gathering explicitly costly rather than trivially dominant.
Reward is delivered only at the final timestep:
\begin{equation}\label{eq:reward}
    R_T(x_T^{(1)}, x_T^{(2)}) = \begin{cases}
        +1.0  & \text{if } x_T^{(1)} = \theta \text{ and } x_T^{(2)} = 4\,,                     \\
        -1.0  & \text{if } x_T^{(1)} \neq \theta,\; x_T^{(1)} < 5 \text{ and } x_T^{(2)} = 4\,, \\
        +0.33 & \text{if } x_T^{(1)} = 5\,,                                                     \\
        -0.33 & \text{if } x_T^{(1)} \neq 5 \text{ and } x_T^{(2)} < 4\,.
    \end{cases}
\end{equation}
Maximum reward ($+1.0$) therefore requires high reactivity, cue use, and correct navigation, whereas reducing the knob guarantees the safe sink ($+0.33$) without either epistemic behavior or state-contingent control. The $-0.33$ outcome for low-reactivity non-safe endings keeps these partially committed trajectories worse than the safe fallback.

% -----------------------------------------------------------------------------
\subsection{Objectives and Posterior Constraints}\label{sec:three-objectives}
% -----------------------------------------------------------------------------

For all three objectives, inference is performed over the same temporally factorized joint posterior,
\begin{equation}\label{eq:temporal-factorization}
    q(\bm{y}, \bm{x}, \bm{u}, \theta) = q(\theta) \prod_{t=1}^{T} q(u_t \mid x_{t-1}, \theta)\, q(x_t \mid x_{t-1}, u_t, \theta)\, q(y_t \mid x_t, \theta)\,,
\end{equation}
which is a structured approximation used for scalability in the experiments.
This parameterization makes the state- and context-conditioned conditional marginals explicit, but it is not part of the conceptual claim from \capsecref{sec:methodology}; it is the experimental posterior family used to instantiate that claim.
Conditioning on $\theta$ does not give the executed action privileged access to the latent context: the first action marginalizes over the current belief $q(\theta)$, while future $\theta$-dependence represents branches induced by possible cue observations.

To isolate the contribution of epistemic priors, we define three Variational Free Energy objectives over the same joint posterior $q(\bm{y}, \bm{x}, \bm{u}, \theta)$:
\begin{align}
    F_{\text{marginal}}[q] & = \Ex{q(\bm{y}, \bm{x}, \bm{u}, \theta)}{\log \frac{q(\bm{y}, \bm{x}, \bm{u}, \theta)}{p(\bm{y}, \bm{x}, \bm{u}, \theta)\, \hat{p}(\bm{x})}}\,, \label{eq:F-marginal}                                                                   \\
    F_{\text{planning}}[q] & = F_{\text{marginal}}[q] + \sum_{t=1}^{T} \left(\ent{q(x_{t-1}, u_t)} - \ent{q(x_{t-1})}\right)\,, \label{eq:F-planning}                                                                                                                \\
    F_{\text{active}}[q]   & = \Ex{q(\bm{y}, \bm{x}, \bm{u}, \theta)}{\log \frac{q(\bm{y}, \bm{x}, \bm{u}, \theta)}{p(\bm{y}, \bm{x}, \bm{u}, \theta)\, \hat{p}(\bm{x})\, \tilde{p}(\bm{x})\, \tilde{p}(\bm{u})\, \tilde{p}(\bm{y}, \bm{x})}}\,. \label{eq:F-active}
\end{align}
All three objectives use the temporal factorization from \eqref{eq:temporal-factorization} and therefore share the same capacity to represent inner-horizon closed-loop control.
They differ in the objective terms they optimize:
\begin{itemize}[nosep]
    \item $F_{\text{marginal}}$ includes only the generative model and preference prior $\hat{p}(\bm{x})$.
          It serves as a baseline with no epistemic drive.
    \item $F_{\text{planning}}$ adds a correction term that accounts for the non-manipulability of environment noise \cite{lazaro-gredilla_what_2024}, addressing optimistic inference.
          Intuitively, this prevents the optimizer from taking credit for uncertainty reduction that comes only from exogenous transition noise rather than from controllable action consequences.
          It has no epistemic drive but produces correct instrumental behavior.
    \item $F_{\text{active}}$ includes the full set of epistemic priors from \eqref{eq:epistemic-priors}, thereby incentivizing information-seeking within the same state-action posterior family.
\end{itemize}
Comparing these three objectives on the same environment and with the same inference procedure isolates the contribution of the epistemic objective while holding the state-action posterior family fixed.

To test that representational claim directly, we also evaluate an action-state factorized variant of $F_{\text{active}}$.
This variant keeps the active objective and its epistemic priors, but constrains the posterior so that future actions are independent of future states and the latent context:
\begin{equation}\label{eq:action-state-factorized-posterior}
    q_{\perp}(\bm{y}, \bm{x}, \bm{u}, \theta)
    =
    q(\bm{y}, \bm{x}, \theta) \prod_{t=1}^{T} q(u_t)\,.
\end{equation}
This is not a full mean-field posterior: the future states may still be represented jointly with each other and with $\theta$.
The constraint specifically removes the dependence needed to form state- and context-conditioned action distributions, while preserving the epistemic drive that makes the cue valuable.

% -----------------------------------------------------------------------------
\subsection{Inference Procedure}\label{sec:inference-procedure}
% -----------------------------------------------------------------------------

Planning is performed by minimizing the Variational Free Energy over a joint posterior with the temporal factorization from \eqref{eq:temporal-factorization}, using the Adam optimizer~\cite{kingma_adam_2015} implemented in JAX~\cite{bradbury_jax_2018}.
At each environment step, the agent receives an observation and updates its belief over the current state and $\theta$.
The agent then reindexes the prospective rollout from the current belief state $x_0$, minimizes the VFE objective over a planning horizon of $T = 3$ for 1000 optimization steps, and executes the first rollout action with highest marginal probability:
\begin{equation}\label{eq:action-selection}
    \hat{u}_1 = \arg\max_{u_1} q(u_1)\,.
\end{equation}
Here $q(u_1)$ denotes the current marginal over the first action in the newly indexed rollout, induced by the joint posterior after conditioning on the agent's present beliefs.
Thus the environment advances in real time, but the planning variables are always indexed locally as $x_{0:T}$ and $u_{1:T}$.
This receding-horizon scheme replans at each timestep, incorporating new observations.
Each episode runs for a maximum of 5 environment steps.
The same optimizer and receding-horizon procedure apply to all VFE conditions from \capsecref{sec:three-objectives}.
For the three state-action joint posterior objectives, only the objective function differs between conditions; for the factorized active condition, the active objective is held fixed while the posterior constraint in \eqref{eq:action-state-factorized-posterior} is imposed.

% -----------------------------------------------------------------------------
\subsection{Experimental Setup}\label{sec:experimental-setup}
% -----------------------------------------------------------------------------

Six methods are compared: the three state-action joint posterior VFE objectives above, the action-state factorized active objective, Sophisticated Inference, and Standard Expected Free Energy planning~\cite{friston_sophisticated_2021} implemented in pymdp~\cite{heins_pymdp_2022}.
Both pymdp baselines use $T=3$, deterministic action selection, $\gamma=16$, utility, and state/parameter information gain; SI uses one-step policies with tree-search horizon $T$, while Standard EFE evaluates length-$T$ action sequences without SI tree search.
Together they span epistemic drive (present or absent) crossed with action representation (state-action joint, action-state factorized, or open-loop sequence based).

Each method is evaluated over 100 episodes with $\theta$ sampled uniformly per episode. \footnote{Experiment code is available at \url{https://github.com/biaslab/EFEPolicyCoordinate}.}
Four metrics are reported: mean reward, optimal rate (reached the correct goal with full reactivity), safe rate (reached the safe sink), and cue visit rate.
The cue visit rate is the most diagnostic metric, as it directly reveals whether a method values information-gathering.
All reported percentages are empirical rates over those 100 episodes, and we report exact 95\% Clopper--Pearson confidence intervals for the Bernoulli rate metrics.
Among the three state-action joint posterior VFE objectives, the posterior family is held fixed, so differences isolate the effect of epistemic priors.
The comparison between $F_{\text{active}}$ and its action-state factorized variant instead holds the active objective fixed and isolates whether posterior dependencies between actions, states, and context are necessary.
Across the full six-method comparison, the contrast between open-loop, factorized, and state-action joint methods tests whether contingent action selection is necessary to exploit the high-reward path.

\FloatBarrier
% =============================================================================
\section{Results}\label{sec:results}
% =============================================================================

\begin{table}[t]
    \centering
    \small
    \caption{Performance on the Reactivity Maze over 100 episodes. Optimal rate denotes the fraction of episodes achieving terminal reward $+1.0$, i.e., reaching the correct goal with full reactivity. Confidence intervals for rate metrics are exact 95\% intervals.}
    \label{tab:reactivity-maze-results}
    \begin{tabular}{lcccc}
        \toprule
        Method                          & Mean reward    & Optimal rate           & Safe rate      & Cue visit rate          \\
        \midrule
        $F_{\text{marginal}}$           & 0.330          & 0\% [0,4]              & 100\% [96,100] & 0\% [0,4]               \\
        $F_{\text{planning}}$           & -0.120         & 44\% [34,54]           & 0\% [0,4]      & 0\% [0,4]               \\
        $F_{\text{active}}$             & \textbf{0.967} & \textbf{98\%} [93,100] & 0\% [0,4]      & \textbf{100\%} [96,100] \\
        \addlinespace
        $F_{\text{active}}^{u \perp x}$ & -0.060         & 1\% [0,5]              & 43\% [33,53]   & \textbf{100\%} [96,100] \\
        Standard EFE planning           & 0.065          & 12\% [6,20]            & 46\% [36,56]   & 55\% [45,65]            \\
        \addlinespace
        Sophisticated Inference         & 0.927          & 95\% [89,98]           & 0\% [0,4]      & \textbf{100\%} [96,100] \\
        \bottomrule
    \end{tabular}
\end{table}

The results in \autoref{tab:reactivity-maze-results} follow the two-axis analysis from \capsecref{sec:experimental-setup}. First, holding the state-action joint posterior fixed shows the role of the objective. The two objectives without epistemic priors do not value the cue: $F_{\text{marginal}}$ converges to the safe solution in every episode, while $F_{\text{planning}}$ remains instrumentally aggressive but never visits the cue and reaches the correct goal only $44\%$ of the time, close to chance for the binary goal without the disambiguating cue. Adding epistemic priors changes this pattern. $F_{\text{active}}$ visits the cue in every episode and achieves the optimal outcome $98\%$ of the time, with the remaining failures appearing episodic rather than systematic cue avoidance.

The next two rows test the complementary question: what happens when epistemic value is present, but the representation cannot fully express state- and context-contingent action selection. The action-state factorized active variant keeps the same active objective as $F_{\text{active}}$, and it also visits the cue in every episode. Its optimal rate nevertheless falls to $1\%$, showing that information gathering is not enough when the posterior cannot propagate state- and context-conditioned future goals back into present action selection. Standard EFE planning shows the analogous limitation for open-loop sequence evaluation: it contains epistemic value, but evaluates fixed action sequences, and therefore reaches the optimal outcome in only $12\%$ of episodes.

The successful methods are the ones that combine both ingredients. Sophisticated Inference reaches the optimal outcome $95\%$ of the time, closely matching $F_{\text{active}}$ within the reported confidence intervals. This supports the main claim of this work: the advantage usually attributed to Sophisticated Inference is not uniquely tied to tree search itself, but to combining epistemic value with an inner-horizon closed-loop representation. That the two methods use different inference procedures, tree search for Sophisticated Inference and gradient-based VFE minimization for $F_{\text{active}}$, is by design: because the claim is behavioral and representational, recovering the same regime through a different procedure is the intended result rather than a confound.

Optimization diagnostics are reported in \refappx{app:convergence}.
All three VFE objectives converge reliably under the experimental optimizer; the difference between them is the behavior they converge to, not whether optimization converges.

\section{Discussion and Conclusion}\label{sec:discussion}

The Reactivity Maze separates epistemic value from inner-horizon closed-loop control. The cue is useful only if the planner can represent future actions as conditional on the state and context revealed after visiting it. The results support that decomposition: methods without epistemic drive do not seek the cue, methods that remove future action dependence on state and context do not fully exploit it, and only $F_{\text{active}}$ and Sophisticated Inference reliably recover the intended information-sensitive behavior.

The decisive ablation is $F_{\text{active}}^{u \perp x}$: it preserves the active inference objective and visits the cue in every episode, but performs similarly to Standard EFE planning because the posterior constraint prevents action choices from depending on the future states and contexts revealed by that cue.
Thus the failure is not a failure to update beliefs across real time, but a failure to represent conditional future control within the rollout.

This sharpens the interpretation of Sophisticated Inference. The contribution is not to rediscover closed-loop control, but to show that the closed-loop structure implemented by SI can be represented in epistemic-prior VFE minimization when the variational family permits conditionals such as $q(u_t\mid x_{t-1}, \theta)$. In this benchmark, epistemic-prior active inference reaches the same qualitative regime as SI by minimizing VFE over a joint posterior that can express state- and context-conditioned actions. The point of the comparison is therefore not that tree search is unnecessary in every setting, but that the behavior associated with sophistication can be explained by the objective and the posterior dependencies rather than by a particular planning procedure.

Seen this way, the present experiment is best understood as an ablation study on sophistication. It isolates what must be present for active inference agents to exploit informative cues under stochastic dynamics, while comparing posterior families that preserve or remove future action dependence on states and context. Questions of large-scale computational realization are important, but they are downstream of this representational point. The advantage we identify is therefore conceptual rather than computational: reproducing the behavior associated with Sophisticated Inference does not require a dedicated planning procedure, only variational free-energy minimization over a joint posterior, which is the inference already used throughout active inference. Whether this also yields a computational or scaling advantage is a separate question that we do not take up here. In particular, \cite{nuijten_message_2026} shows that epistemic-prior message passing can be realized in larger state spaces; the present paper identifies why that route should recover the same behavioral advantage often credited to Sophisticated Inference.

In short, epistemic priors supply the active-inference objective, while a joint future posterior supplies the inner-horizon closed-loop control structure. Tree search and VFE minimization are different realizations of that structure, provided the latter does not factor future actions away from the future variables that make action selection contingent.

\begin{credits}
\subsubsection{\ackname}
\begin{sloppypar}
This publication is part of the project ROBUST: Trustworthy AI-based Systems for Sustainable Growth with project number KICH3.LTP.20.006, which is (partly) financed by the Dutch Research Council (NWO), GN Hearing, and the Dutch Ministry of Economic Affairs and Climate Policy (EZK) under the program LTP KIC 2020-2023.
\end{sloppypar}

\subsubsection{\discintname}
The authors have no competing interests to declare.
\end{credits}

\bibliographystyle{splncs04}
\bibliography{references}

\clearpage

\appendix
% =============================================================================
\renewcommand{\thesection}{A}
\renewcommand{\theHsection}{appendix.\thesection}
\section{Convergence Diagnostics}\label{app:convergence}
% =============================================================================

This appendix reports the optimization diagnostics underlying the experiments in \capsecref{sec:results}.
We include aggregate convergence summaries and representative scenario-level trajectories to show that the reported behavioral differences are driven by the objectives themselves rather than by unstable optimization.

\begin{figure}[h]
    \centering
    \resizebox{\linewidth}{!}{\import{figures/convergence/}{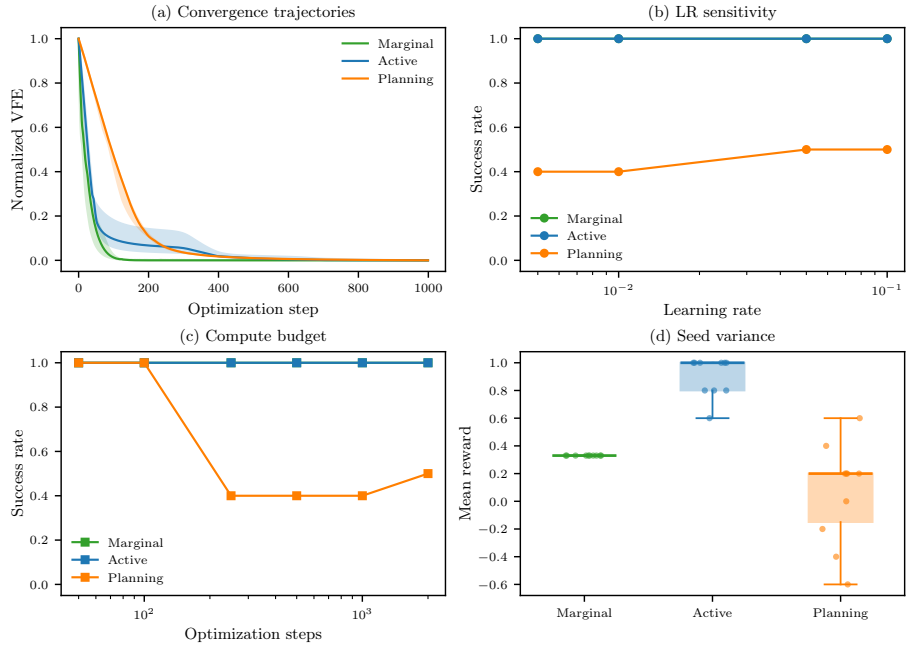}}
    \caption{Optimization diagnostics for the three VFE objectives. Panel (a) shows normalized loss trajectories; panel (b) shows sensitivity to learning rate; panel (c) shows sensitivity to optimization budget; panel (d) shows variation across random seeds. Across all settings, the objectives converge reliably. The main behavioral differences therefore reflect differences in the objective, not failures of optimization.}
    \label{fig:convergence}
\end{figure}

\begin{figure}[h]
    \centering
    \resizebox{\linewidth}{!}{\import{figures/convergence/}{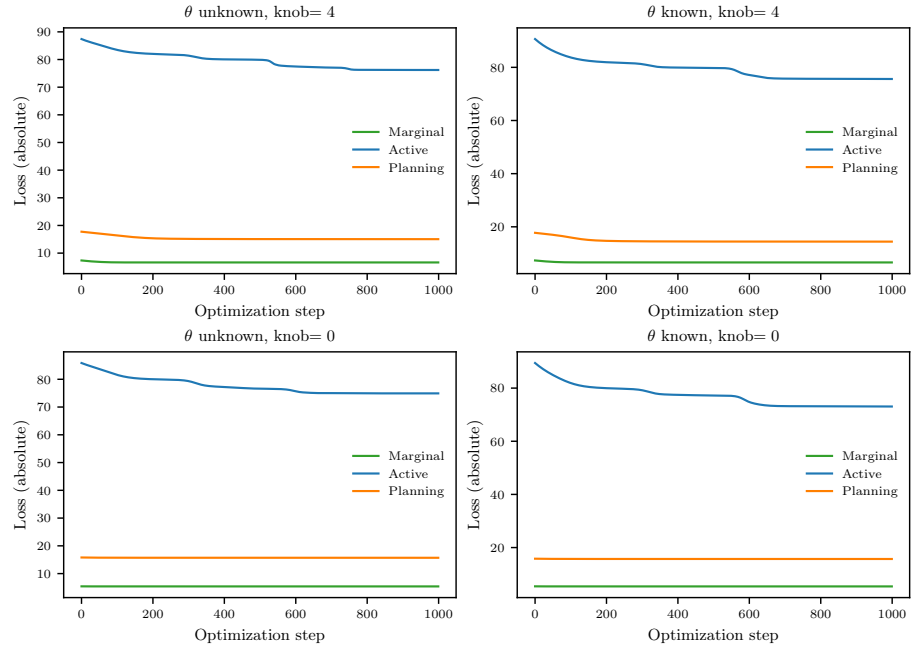}}
    \caption{Loss curves for four representative planning scenarios. The qualitative convergence pattern is consistent across known and unknown goal contexts and across low- and high-reactivity settings.}
    \label{fig:scenario-convergence}
\end{figure}

The optimization diagnostics confirm that the empirical differences in \capsecref{sec:results} are not artifacts of poor convergence.
$F_{\text{marginal}}$, $F_{\text{planning}}$, and $F_{\text{active}}$ all show smooth loss reduction across optimization steps, and their qualitative behavior is preserved across the tested learning rates and compute budgets.

The seed analysis shows the same pattern at the level of outcomes.
$F_{\text{marginal}}$ consistently converges to the safe solution, $F_{\text{active}}$ consistently converges to a cue-seeking high-reward policy, and $F_{\text{planning}}$ consistently converges to a cue-averse but non-safe policy.
The important point is that all three objectives are stable in the optimization sense.
What differs is which policy each objective makes attractive.

\end{document}